\documentclass[10pt,twocolumn,letterpaper]{article}

\usepackage{cvpr}              % To produce the CAMERA-READY version
\makeatletter
\@namedef{ver@everyshi.sty}{}
\makeatother

\usepackage{graphicx}
\usepackage{amsmath}
\usepackage{amssymb}
\usepackage{booktabs}
\usepackage{svg}
\usepackage{multicol}
\usepackage{multirow}

\definecolor{cvprblue}{rgb}{0.21,0.49,0.74}
\usepackage[pagebackref,breaklinks,colorlinks,allcolors=cvprblue]{hyperref}

\def\paperID{2090} % *** Enter the Paper ID here
\def\confName{CVPR}
\def\confYear{2025}

\title{TexGaussian: Generating High-quality PBR Material via \\ Octree-based 3D Gaussian Splatting}

\author{Bojun Xiong$^{1}$\thanks{Denotes equal contribution.}~\thanks{This work was partly done when Bojun and Jiakui interned in Baidu.}~, Jialun Liu$^{2}$\footnotemark[1]~, Jiakui Hu$^{3}$\footnotemark[2]~, Chenming Wu$^{2}$, Jinbo Wu$^{2}$, Xing Liu$^{2}$, \\
Chen Zhao$^{2}$, Errui Ding$^{2}$, Zhouhui Lian$^{1}$\thanks{Corresponding author. E-mail: lianzhouhui@pku.edu.cn\\This work was supported by the National Natural Science Foundation of China (Grant No.: 62372015), Center For Chinese Font Design and Research, and Key Laboratory of Intelligent Press Media Technology.} \vspace{6pt}\\
    $^1$Wangxuan Institute of Computer Technology, Peking University, China\\
    $^2$Baidu VIS \\
    $^3$Institute of Medical Technology, Peking University, China\\
}

\makeatletter
\let\@oldmaketitle\@maketitle%
\renewcommand{\@maketitle}{\@oldmaketitle%
 \centering
 \vspace{-5mm}
    \includegraphics[width=\textwidth]{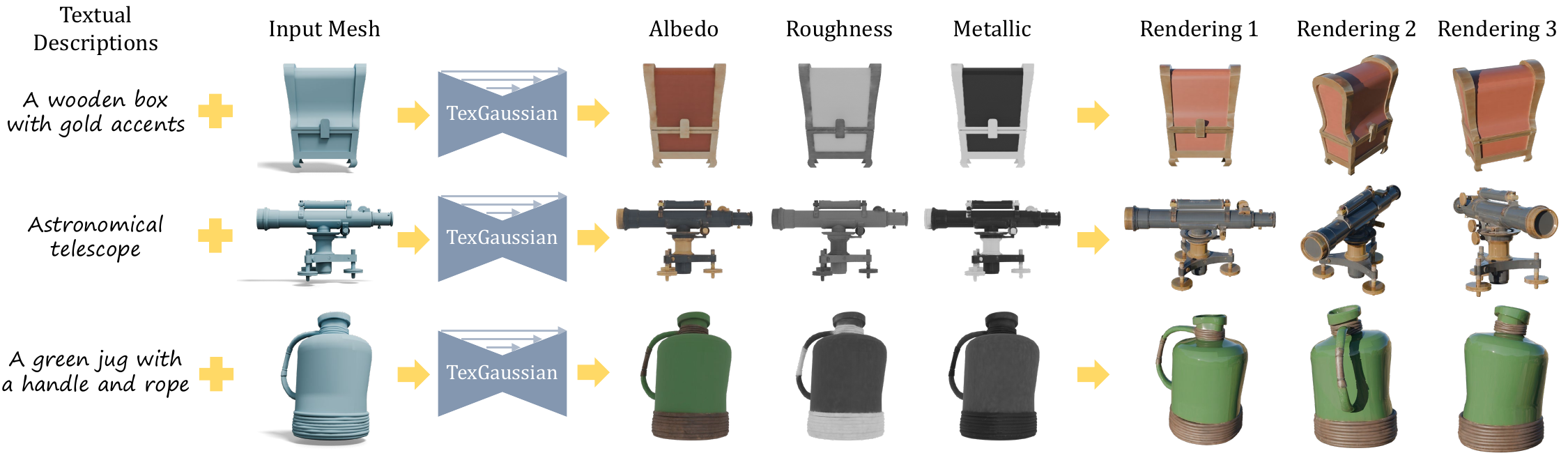} 
     \captionof{figure}{Our proposed TexGaussian is capable of generating high-quality materials for given input 3D meshes based on the corresponding textual descriptions. The generated results are naturally compatible with modern graphical engines for photo-realistic rendering under different environment maps.}
    \label{fig:teaser}
    \bigskip}
\makeatother

\begin{document}
\maketitle

\begin{abstract}
Physically Based Rendering (PBR) materials play a crucial role in modern graphics, enabling photorealistic rendering across diverse environment maps. Developing an effective and efficient algorithm that is capable of automatically generating high-quality PBR materials rather than RGB texture for 3D meshes can significantly streamline the 3D content creation. Most existing methods leverage pre-trained 2D diffusion models for multi-view image synthesis, which often leads to severe inconsistency between the generated textures and input 3D meshes. This paper presents TexGaussian, a novel method that uses octant-aligned 3D Gaussian Splatting for rapid PBR material generation. Specifically, we place each 3D Gaussian on the finest leaf node of the octree built from the input 3D mesh to render the multi-view images not only for the albedo map but also for roughness and metallic. Moreover, our model is trained in a regression manner instead of diffusion denoising, capable of generating the PBR material for a 3D mesh in a single feed-forward process. Extensive experiments on publicly available benchmarks demonstrate that our method synthesizes more visually pleasing PBR materials and runs faster than previous methods in both unconditional and text-conditional scenarios, exhibiting better consistency with the given geometry. Our code and trained models are available at \url{https://3d-aigc.github.io/TexGaussian}.

\end{abstract}

%%%%%%%%% BODY TEXT
\vspace{-4mm}
\section{Introduction}
\label{sec:intro}

Traditional 3D asset creation relies heavily on the expertise and extensive effort of professional designers~\cite{labschutz2011content}, posing a significant barrier for casual users interested in creating 3D models independently. In the 3D design process, geometry creation typically represents only a small portion of the overall time, while the majority is dedicated to developing textures and appearances, which are far more time-consuming. Achieving a delicate appearance for a 3D model often demands substantial time and effort from experienced designers.

Recently, Artificial Intelligence Generated Content (AIGC) based on deep generative models, especially diffusion models~\cite{sohl2015deep, ho2020denoising} have been widely used to facilitate the process of artistic creation, catalyzing advancements in image generation~\cite{podell2024sdxl, ramesh2022hierarchical, rombach2021highresolution, saharia2022photorealistic, dhariwal2021diffusion} and video generation~\cite{ho2022video, blattmann2023align, blattmann2023stable, videoworldsimulators2024, xing2024tooncrafter}. As a result, exploring effective ways to leverage deep generative models to streamline the creation of detailed appearances for 3D models has become a popular direction in the graphics and vision communities.

% as well as 3D model synthesis ~\cite{poole2022dreamfusion, zhang2024clay, wang2023prolificdreamer}. 

Recent advancements in 3D texture generation attempt to use depth-conditional pre-trained 2D diffusion model~\cite{zhang2023adding} to synthesize RGB images based on the depth maps rendered from multiple views, such as TEXture~\cite{richardson2023texture} and text2tex~\cite{chen2023text2tex}. Subsequent works~\cite{cao2023texfusion, huo2024texgen, zhang2024texpainter} further improve multi-view consistency via blending the multi-view images to a single and consistent texture map in every denoising step. However, these methods struggle to have a global picture of 3D geometries due to the use of 2D diffusion models, leading to inconsistencies between the texture map and the semantics of the input 3D meshes. Moreover, the generated assets suffer from illumination-baked textures, which can significantly degrade the quality of the final rendering when placed in novel lighting conditions~\cite{zeng2024paint3d}. While DreamMat~\cite{zhang2024dreammat} supplements geometry and light control to achieve material decomposition through score distillation sampling~\cite{poole2022dreamfusion}, it still struggles to fully capture the global geometry features. This limitation often results in the multi-face Janus problem and leads to over-saturated colors.

% Although these methods are capable of synthesizing diverse RGB texture for input 3D mesh owing to the powerful 2D diffusion prior~\cite{rombach2021highresolution}, they can not guarantee the multi-view consistency due to the error accumulation stemming from the autoregressive view inpainting process. In addition, the generated results have issues with pre-illuminated textures, which would damage the final rendering results in new illumination environments~\cite{zeng2024paint3d}. To improve multi-view consistency, some subsequent methods~\cite{cao2023texfusion, huo2024texgen, zhang2024texpainter} proposed to blend the multi-view image to a single and consistent texture map in every denoising steps. To achieve results with decoupled lighting and textures, DreamMat~\cite{zhang2024dreammat} added geometry and light conditions with ControlNet~\cite{zhang2023adding} to realize material decomposition. However, all these methods struggle to understand 3D geometries globally due to the absence of a 3D neural network, leading to global inconsistencies between the texture map and the semantics of the input 3D models.

On the other hand, training a 3D neural network directly on 3D data, such as Point-UV Diffusion~\cite{yu2023texture} and TexOct~\cite{Liu_2024_CVPR}, is an effective way for 3D global consistency. Meanwhile, this avoids multi-view sampling and score distillation sampling, which accelerates the process of texture synthesis. However, relying on colored point clouds for 3D representation and supervision often results in blurred outputs, primarily due to the sparse and non-compact nature of point clouds in 3D space. Due to the limitations of the adopted 3D representations and the lack of training data on PBR materials, these approaches are incapable of generating high-fidelity PBR materials for 3D models. 

To address the aforementioned challenges, this paper presents TexGaussian, a fast and high-fidelity PBR material generation model directly in 3D space that maintains 3D global consistency. Different from previous approaches that primarily rely on diffusion models, our method works in a regression manner to regress the PBR material from the input mesh for faster generation speed. To enable effective learning in 3D space, we propose to use octree, a specialized sparse voxel structure that efficiently organizes and preserves 3D information, which can be built from 3D point clouds sampled from the surface of the object. However, directly regressing the color of 3D point clouds on octree often results in blurry textures as mentioned in~\cite{Liu_2024_CVPR}. To tackle the challenges of the incompact and discrete nature presented by points, we use 3D Gaussian Splatting (3DGS)~\cite{kerbl3Dgaussians}, a robust representation that bridges the gap between 3D space and 2D raster images, allowing us to fully utilize rich 2D image information to alleviate blurring results. Specifically, for each input mesh, we sample the dense 3D point clouds on its surface to build the corresponding octree. On each octant (i.e., the finest leaf node of octree), we place a 3D Gaussian~\cite{kerbl3Dgaussians} at its central position. Then, we use the octree-based 3D U-Net~\cite{Wang-2017-ocnn} to predict the parameters of each 3D Gaussian on octants. Apart from RGB colors, we extend each 3D Gaussian with additional parameters to represent the roughness and metallicity of 3D objects. Multi-view images, including albedo, roughness, and metallic maps, can be rasterized from all these 3D Gaussians via 3DGS. The 3D U-Net is supervised by the difference between the predicted multi-view images and their corresponding ground truth. Notably, the 3D U-Net is trained to directly regress the multi-view images based on the geometry feature of the input 3D model, which further facilitates the process of 3D Gaussian prediction compared to diffusion manner. We train our TexGaussian model on a subset of Objaverse~\cite{deitke2023objaverse} with high-quality PBR materials, enabling fast PBR material generation with a single feed-forward pass. In summary, the contributions of our paper are threefold:

\begin{itemize}
\item We propose an octant-aligned 3D Gaussian Splatting method for high-quality PBR material synthesis on untextured input 3D mesh, which fully utilizes the supervision from 2D images, avoiding blurry results caused by the discreteness of 3D point clouds.
\item We adopt a regression manner to train our 3D U-Net model instead of diffusion denoising, achieving faster generative speed.

\item We propose TexGaussian, a novel PBR material generation method based on the above two techniques. To our knowledge, our work is the first to generate PBR materials directly in 3D space. Qualitative and quantitative experiments have been conducted to verify the superiority of the quality and efficiency of our method over other existing approaches.
\end{itemize}

\vspace{-2mm}
\section{Related Work}
In this section, we mainly summarize current texture synthesis methods, which can be roughly divided into three categories.

\subsection{Multi-view Images Synthesis}
% With the rapid development of large-scale Text-to-Image (T2I) models~\cite{rombach2021highresolution, ramesh2022hierarchical, saharia2022photorealistic}, 
Many previous works have tried to leverage the powerful T2I model to assist texture generation for 3D shapes. Specifically, they render the depth map of input 3D mesh from multiple views and use depth conditional T2I models~\cite{zhang2023adding} to synthesize RGB images and perform text-conditioned texture synthesis. TEXTure~\cite{richardson2023texture} and Text2Tex~\cite{chen2023text2tex} iteratively paint a mesh from different views. However, images synthesized in early view could produce errors that are not reconcilable with the geometry that is observed in later views. Many subsequent works try to alleviate multi-view inconsistency via different alignment modules. TexFusion~\cite{cao2023texfusion} proposes a sequential interlaced multi-view sampler that interleaves texture assembling with denoising steps in different camera views. Similarly, TexGen~\cite{huo2024texgen} directly enforces view consistent sampling in RGB texture space and develops a noise resampling strategy to retain rich texture details. TexPainter~\cite{zhang2024texpainter} blends images from different views into a common color-space texture image by weighted averaging to guarantee multi-view consistency. To remove light influence from 2D diffusion models, Paint3D~\cite{zeng2024paint3d} contribute separate UV Inpainting and UVHD diffusion models specialized in shape-aware refinement. More recently, Meta 3D TextureGen~\cite{bensadoun2024meta} utilizes geometry-aware text-to-image model to generate a multi-view image for texture. Although these methods achieve impressive texture results, they can still hardly comprehend the overall geometry of input 3D mesh.

\subsection{Optimization-based 3D Generation}
Before the emergence of large-scale Text-to-Image generative model, earlier methods~\cite{hong2022avatarclip, chen2022tango, ma2023x, michel2022text2mesh, mohammad2022clip} propose to optimize texture map of 3D object via natural language supervised visual model, CLIP~\cite{radford2021learning}. Subsequently, score distillation sampling (SDS) was adopted by DreamFusion~\cite{poole2022dreamfusion} and Magic3D~\cite{lin2023magic3d}. The key idea is to optimize 3D representations such as NeRF~\cite{mildenhall2020nerf} or InstantNGP~\cite{mueller2022instant} with the gradient guidance from 2D diffusion priors~\cite{poole2022dreamfusion, wang2023prolificdreamer}. To generate PBR material, TextureDreamer~\cite{yeh2024texturedreamer} optimizes spatially-varying bidirectional reflectance distribution (BRDF) field through personalized geometric-aware score distillation. Fantasia3D~\cite{chen2023fantasia3d} uses a single predefined environmental. However, the generated images from diffusion models may not be consistent with the given environment light. Another concurrent work Paint-it~\cite{youwang2024paintit} is similar to Fantasia3D~\cite{chen2023fantasia3d} but with Deep Convolutional Physically-Based Rendering (DC-PBR) parameterization. DreamMat~\cite{zhang2024dreammat} proposes a novel geometry and light-aware diffusion model, which is trained to generate images that are consistent with the given environment light. FlashTex\cite{deng2024flashtex} uses a light-conditioned diffusion model within a two-stage pipeline, combining reconstruction and SDS optimization to enhance texture quality and achieve better light disentanglement. However, these methods struggle with the Janus problem due to the semantically ambiguous. And the time consumption is relatively too long to use in practice.

\subsection{Generating Texture from 3D Data}
The most straightforward way to synthesize texture map for 3D mesh is to train generative model directly from 3D data with texture groud truth~\cite{collins2022abo, shapenet2015, deitke2023objaverse, objaverseXL}. Early methods such as Texture Fields~\cite{OechsleICCV2019} learn implicit texture fields to assign a color to each pixel on the surface of the 3D shape. Texturify~\cite{siddiqui2022texturify} devices face convolution operation on mesh surface to predict texture on each face. It employs differentiable rendering with an adversarial loss to ensure that generated textures produce realistic imagery. Recently, some diffusion-based texture synthesis methods, such as Point-UV~\cite{yu2023texture} and TexOct~\cite{Liu_2024_CVPR} train a denoising network on colors of point clouds which are further mapped to 2D UV map. Although these methods achieve better 3D global consistency with input mesh, they are only trained on several categories of small datasets~\cite {shapenet2015}. What's more, discrete supervision from 3D point clouds leads to suboptimal results compared with continuous signals such as 2D images.

\begin{figure*}[t!]
  \centering
  \includegraphics[width=\textwidth]{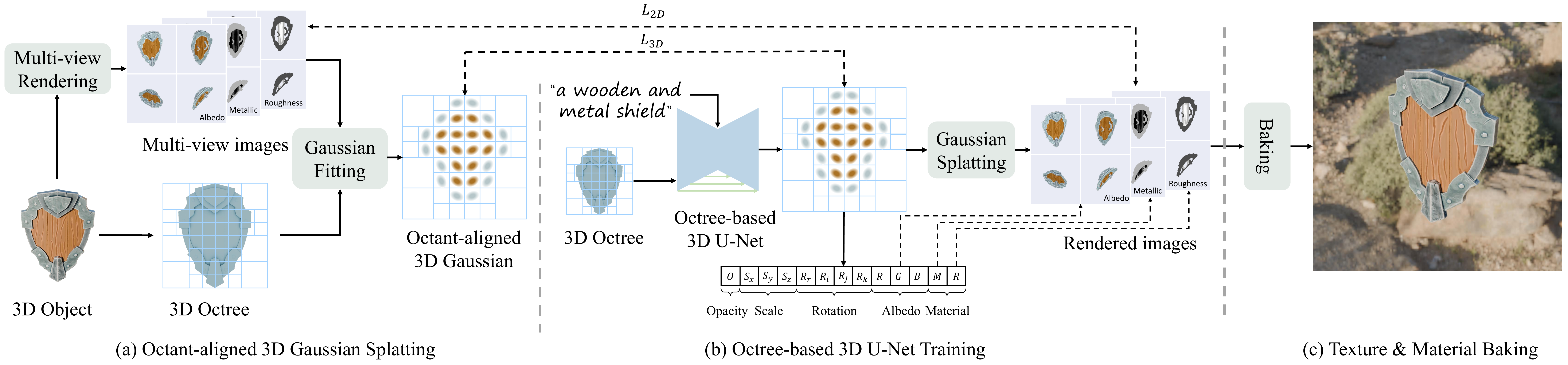}
  \caption{An overview of our PBR material generation framework. (a) We propose octant-aligned 3D Gaussian Splatting, which positions a 3D Gaussian at the center of each finest leaf node of the constructed octree. Additional channels are added at the end of the Gaussian parameters to model PBR material. (b) We use the 3D U-Net built upon octree-based convolutional networks to predict the Gaussian parameters. Our octree-based 3D U-Net is trained by minimizing the difference on 2D raster images and 3D Gaussian parameters. (c) We bake the multi-view rendered images to the UV space of the input 3D model to realize physically based rendering under new illumination environments.}
  \label{fig:method}
\vspace{-2mm}
\end{figure*}

\vspace{-2mm}
\section{Method}
In this section, we provide a detailed explanation of our proposed method, TexGaussian. The overall pipeline of our method is shown in Fig.~\ref{fig:method}. Existing texture synthesis approaches mainly rely on pre-trained 2D diffusion models, which struggle to fully understand the overall 3D structure. This often leads to misalignment between the generated texture map and 3D semantics. Our goal is to synthesize the high-quality PBR material for a given mesh directly in 3D space.

\subsection{Overview}
To enable effective learning in 3D space, we use octree, a sparse voxel structure, to organize and store 3D information without compromising representation quality due to the inherent sparseness of 3D objects in 3D space. Thus, we sample a large number of points on the surface of the given mesh to construct the corresponding octree. The key components of our method are the octant-aligned Gaussian Splatting and the octree-based 3D U-Net. Specifically, we place a 3D Gaussian at the center of each octree’s finest leaf node. The octree-based 3D-Unet is trained to predict the parameters of each 3D Gaussian. Under this circumstance, the generated octant-aligned 3D Gaussians are naturally on the mesh surface. We render them from multiple viewpoints using 3D Gaussian Splatting and train the 3D U-Net by minimizing the difference in 2D raster images and 3D Gaussian parameters. During inference, the rendered multi-view images are baked into the UV space of input mesh using differentiable mesh rendering, producing the final texture and material map. The details of 3D Gaussian Splatting (3DGS) are provided in the supplementary material.

% \subsection{Preliminary}
% Before introducing our method, we first provide a brief background information of 3D Gaussian Splatting. As introduced in~\cite{kerbl3Dgaussians}, Gaussian splatting employs a collection of 3D Gaussians to represent 3D data. Specifically, each Gaussian is defined by a center $\mathbf{x} \in \mathbb R^3$, a scaling factor $\mathbf{s} \in \mathbb R^3$, and a rotation quaternion $\mathbf{q} \in \mathbb R^4$. Additionally, an opacity value $\alpha \in \mathbb R$ and a color feature $\mathbf{c} \in \mathbb R^C$ are maintained for rendering, where spherical harmonics can be used to model view-dependent effects. These parameters can be collectively denoted by ${\Theta}$, with ${\Theta}_i = \{\mathbf{x}_i, \mathbf{s}_i, \mathbf{q}_i, \alpha_i, \mathbf{c}_i\}$ representing the parameters for the $i$-th Gaussian. Rendering of the 3D Gaussians involves projecting them onto the image plane as 2D Gaussians and performing alpha blending for each pixel in front-to-back depth order, thereby determining the final color and alpha.

\subsection{Octant-aligned 3D Gaussian Splatting}
For a given mesh, we first sample $N = 100,000$ 3D points on its surface. The corresponding octree is then built by adaptively subdividing the voxels containing those points until the maximum depth is reached. As a result, all of the finest leaf nodes of octree lie along the boundary of the 3D object, which is consistent with the characteristic of the optimized 3D Gaussian. Therefore, we align a 3D Gaussian at the central position of each finest leaf node to effectively model the appearance without compromising the splatting quality. It is worth noting that we do not adjust the position of 3D Gaussians because they are already on the surface of 3D shape and adding additional offset relative to the octant center would not improve the rendering quality through our early experiments. To model PBR material that includes roughness and metallic information, we follow~\cite{jiang2024gaussianshader, shi2023gir} to append two additional channels at the end of 3D Gaussian parameters which are responsible for roughness and metallic map rendering, respectively.

For each 3D object in our dataset, we render multi-view images of albedo, roughness, and metallic maps for the training of 3D Gaussian. The multi-view rendering results of the albedo map are view-independent. So we just use three RGB channels to take the place of the original spherical harmonics. As noted, we exclude the position from 3D Gaussian parameters in our model. Thus, our 3D Gaussian parameters consist of 13 channels in total: one for opacity, three for scale, four for rotation, three for albedo, one for roughness, and one for metallic. To stabilize training, we choose to employ different activation functions compared to the original Gaussian Splatting~\cite{kerbl3Dgaussians}. Specifically, We multiply the softplus-activated scales $\mathbf{s}_i$ with 0.01, ensuring that the initial 3D Gaussians conform to the object’s counter of object at the beginning of training rather than expanding outward.

We pre-fitting the parameters of 3D Gaussians on the constructed octree for each 3D object in our dataset via the original loss function in~\cite{kerbl3Dgaussians} on multi-view albedo images, roughness maps, and metallic maps:

\begin{equation}
    L = (1-\lambda)L_1 + \lambda L_{\text{D-SSIM}},
\end{equation}
where $\lambda = 0.2$.

\subsection{Octree-based 3D U-Net Training}
To handle the encoding of our octant-aligned 3D Gaussian representation, we use the 3D U-Net built upon octree-based convolutional neural networks~\cite{Wang-2017-ocnn} to predict the Gaussian parameters. Inspired by LGM~\cite{tang2024lgm}, the output feature of the 3D U-Net on each octant is treated as the 3D Gaussian parameters, which contain 13 channels, as discussed in the last subsection.

To effectively train our 3D U-Net, we adopt the regressive loss objective, which could further facilitate the generation process. The input to our octree-based 3D U-Net is the geometry feature on each octree's finest leaf node, such as normal and local displacement. For text-conditioned PBR material synthesis, the text feature is extracted by pre-trained CLIP model~\cite{radford2021learning} and is fed to U-Net via the octree-based multi-head cross attention mechanism similar to~\cite{Wang2023OctFormer}. The predicted 3D Gaussians are rasterized from multiple views via Gaussian Splatting~\cite{kerbl3Dgaussians}. At each training step, we rasterize the albedo images, alpha images, roughness and metallic maps from randomly selected eight views. Following~\cite{tang2024lgm}, we apply the mean square error (MSE) loss and the VGG-based LPIPS loss~\cite{zhang2018unreasonable} to the albedo image, roughness map, and metallic map:
\begin{equation}
    L_{\text{albedo}} = L_\text{MSE}(I_{\text{albedo}}, I_{\text{albedo}}^{\text{GT}}) + L_{\text{LPIPS}}(I_{\text{albedo}}, I_{\text{albedo}}^{\text{GT}}),
\end{equation}
\begin{equation}
    L_{\text{R}} = L_{\text{MSE}}(I_{\text{R}}, I_{\text{R}}^{\text{GT}}) + L_{\text{LPIPS}}(I_{\text{R}}, I_{\text{R}}^{\text{GT}}),
\end{equation}
\begin{equation}
    L_{\text{M}} = L_{\text{MSE}}(I_{\text{M}}, I_{\text{M}}^{\text{GT}}) + L_{\text{LPIPS}}(I_{\text{M}}, I_{\text{M}}^{\text{GT}}),
\end{equation}
where `R' and `M' denote roughness and metallic, respectively. We further apply the MSE loss on the alpha image for faster convergence of the shape:
\begin{equation}
    L_{\alpha} = L_{\text{MSE}}(I_{\alpha}, I_{\alpha}^{\text{GT}}).   
\end{equation}

To accelerate the coverage process, we also apply the 3D MSE loss $L_{\text{3D}}$, which calculates the difference between predicted parameters of 3D Gaussians and pre-fitting ones. Finally, the complete loss function of our model is defined
as the sum of all the above losses:
\begin{equation}
    L_\text{total} = L_\text{albedo} + L_\text{R} + L_\text{M} + L_{\alpha} + L_{\text{3D}}.    
\end{equation}

\subsection{Texture and Material Baking}
In the inference stage, we also first build the corresponding octree for the input 3D mesh. Then, we use our trained octree-based 3D U-Net to generate 3D Gaussian on every octant and rasterize them from multiple views. The ultimate output of our method should be a global texture map and material map. Thus, we rasterize the input mesh using the differentiable renderer~\cite{Laine2020diffrast} and optimize its albedo, roughness and metallic parameters via the MSE loss between the Nvdiffrast~\cite{Laine2020diffrast} rendering results and 3D Gaussian rendering results. With adequately optimized implementation, this process takes only about several seconds to bake the multi-view images to an untextured 3D model. After the optimization, our input mesh paired with its albedo and material map is capable of performing physically based rendering in new illumination environments.

\section{Experiments}

\subsection{Implementation Details}
\paragraph{Dataset}
We train our TexGaussian model on two publicly-available datasets: ShapeNet~\cite{shapenet2015} and Objaverse~\cite{deitke2023objaverse}. For 3D objects in the ShapeNet dataset, since they only contain albedo maps without PBR materials, our model is trained to generate RGB textures only. We train our model on four categories of ShapNet: \texttt{bench}, \texttt{car}, \texttt{chair} and \texttt{table} which is consistent with previous works~\cite{yu2023texture, Liu_2024_CVPR}.
We curate a subset of Objaverse models encoded with PBR materials and convert those using the specular-glossiness workflow to the metallic-roughness workflow to achieve a consistent PBR representation. This process resulted in a total of $29,200$ models. For each 3D object, we render its RGBA albedo image, roughness map, and metallic map from 64 views of $512^2$ for training. 

\paragraph{Network Architecture} Our 3D U-Net is built upon octree-based convolutional neural networks~\cite{Wang-2017-ocnn} which only operates on non-empty octree leaf nodes. The depth of our constructed octree is set to 8 (resolution $256^2$). Our 3D U-Net consists of 5 down-sampling and up-sampling blocks. For the text-conditioned generation, cross-attention layers are only inserted at the last two down-sampling blocks, the middle block, and the first two up-sampling blocks. The input channel of our 3D U-Net is set to 4, where 3 for normal vector and 1 for local displacement while the output channel is set to 13 as mentioned above. The resolution of 3D Gaussian Splatting is set to $512^2$. The resolution of the baked albedo and material map is set to $1024^2$.

\vspace{-3mm}
\paragraph{Training Details} We train our model in a single-category and unconditional manner on ShapeNet, i.e., there are 4 TexGaussian models without text conditions in total. This per-category model is trained on 4 NVIDIA A100 (40G) GPUs for about 2 days and is set up for a fair comparison with other existing methods trained on ShapeNet. For the Objaverse dataset, we train a text-conditioned TexGaussian model for PBR material generation on 24 NVIDIA A100 (40G) GPUs for two weeks. For each batch, we randomly sample 8 camera views to calculate loss functions. We adopt the AdamW~\cite{loshchilov2017decoupled} optimizer with a learning rate of $4 \times 10^{-4}$, a weight decay of 0.05, and betas of (0.9,0.95). The learning rate is cosine annealed to 0 during the training and we clip the gradient with a maximum norm of 1.0.

\paragraph{Evaluation metrics}
To effectively assess the quality of our generative results, we adopt the metric proposed by~\cite{zheng2023lasdiffusion}. Specifically, each mesh with the generated PBR material is rendered from 20 uniformly distributed views to get multi-view albedo maps, and PBR rendering results in a new illumination environment. These images are used to calculate the FID~\cite{parmar2021cleanfid} and KID~\cite{bińkowski2018demystifying} scores against those from ground truth to evaluate the quality and diversity of generative PBR materials. For the ShapeNet dataset, we only use albedo maps. The final score is averaged across 20 views, and a lower FID and KID score indicates better generation
quality and diversity. 
% For the text-conditioned TexGaussian model, the semantic alignment between the text prompts and rendered images is quantitatively measured by the CLIP Score, where a higher score indicates a better similarity between the generated appearance and text prompts.

\subsection{Unconditional RGB Texture Generation}
\begin{figure}[t!]
\vspace{-2mm}
  \centering
  \includegraphics[width=0.95\columnwidth]{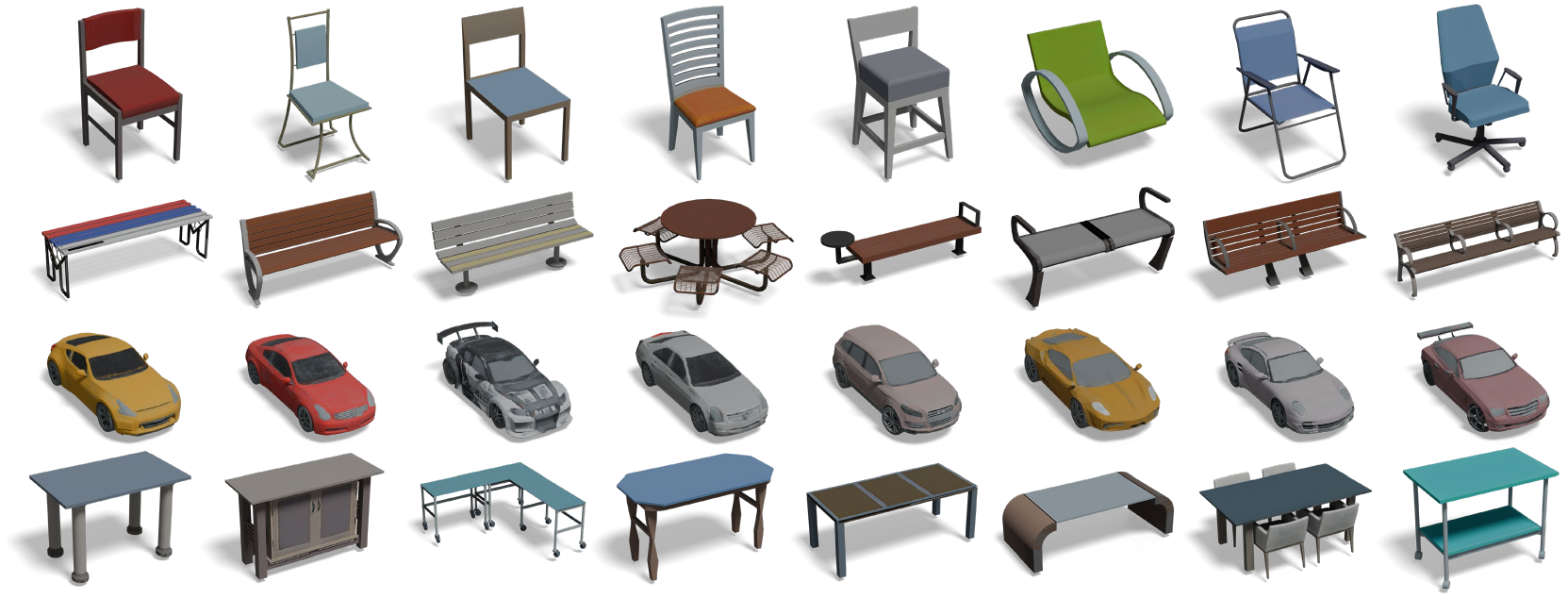}
  \caption{Unconditional RGB texture generative results on ShapeNet. Please zoom in for a better inspection of color details.}
  \vspace{-7mm}
  \label{fig:uncond}
\end{figure}

We conduct unconditional RGB texture instead of PBR material generation on four categories of the ShapeNet dataset. We use the same train and test data split as in~\cite{yu2023texture, Liu_2024_CVPR}. Fig.~\ref{fig:uncond} shows the unconditionally generated RGB texture by our single-category model with high quality, fidelity, and diversity. It can be seen that the texture map synthesized by our method is of great 3D global consistency with the corresponding input geometry. 

\begin{table*}[t]
\centering
\setlength{\tabcolsep}{4.6pt}
\caption{Quantitative comparison of the FID and KID ($\times 10^2$) score as well as the inference time of TexGaussian and other methods on the ShapeNet dataset~\cite{shapenet2015}. The top part reports the comparison with methods that also train per-category models on ShapeNet. The bottom part reports the comparison with 2D diffusion-based methods, which select $50$ samples from the test set of each category.}
\label{tab:uncond}
\begin{tabular}{lccccccccccc}
\toprule
\multirow{2}{*}{Methods} & \multicolumn{2}{c}{Average} & \multicolumn{2}{c}{Bench} & \multicolumn{2}{c}{Car} & \multicolumn{2}{c}{Chair} & \multicolumn{2}{c}{Table} & \multirow{2}{*}{Time}\\ 
\cline{2-11} 
& FID$\downarrow$ & KID$\downarrow$ & FID$\downarrow$ & KID$\downarrow$ & FID$\downarrow$ & KID$\downarrow$ & FID$\downarrow$ & KID$\downarrow$ & FID$\downarrow$ & KID$\downarrow$\\  
\midrule   
% Texturify\cite{siddiqui2022texturify}       & -          & -             & -            & -            & 112.14           & 9.32            & 54.35            & 2.85            & -           & -  & 0.494s  \\ 
Point-UV (1-Stage)~\cite{yu2023texture}       & 88.43          & 5.78             & 64.96            & 1.44            & 186.10           & 17.57            & 46.23            & 1.93            & 56.42           & 2.19   
 & 39.92s    \\ 
Point-UV (2-Stage)~\cite{yu2023texture}       & 61.49          & 2.67         & 67.48            & 1.61           & 89.38           & 5.82            & 41.33            & 1.53            & 47.75           & 1.72    &  49.75s        \\  
TexOct~\cite{Liu_2024_CVPR}       & 59.45          & 2.60        & 60.46            & 0.97           & 90.10           & 6.11            & 37.70            & 1.36            & 49.52           & 1.97   & 17.44s     \\  
Ours    & \textbf{49.76}       & \textbf{2.07}            & \textbf{46.37}   & \textbf{0.44}   & \textbf{80.20}       & \textbf{5.22}       & \textbf{29.96}       & \textbf{1.09}       & \textbf{42.52}  & \textbf{1.54}   & \textbf{11.02s}  \\ 
\midrule
TEXTure~\cite{richardson2023texture}       & 169.82          &  7.94       & 152.53           & 5.28          & 236.58          & 18.52            & 119.71            & 2.86           & 170.46           & 5.11   & 132.21s        \\  
Text2Tex~\cite{chen2023text2tex}       & 156.62          &  6.68       & 154.95           & 5.79         & 188.60          & 15.06            & 125.34           & 3.11           & 157.57           & 2.77  &   1005.58s \\ 
Paint3D~\cite{zeng2024paint3d}       & 135.25          &  4.67       & 104.16           & 1.68         & 204.22          & 13.06            & 104.64           & 2.24           & 127.98           & 1.71 &   231.56s \\ 
Ours    & \textbf{100.97}       & \textbf{1.88}            & \textbf{86.37}   & \textbf{0.75}   & \textbf{117.53}       & \textbf{5.20}       & \textbf{94.89}       & \textbf{0.57}       & \textbf{105.09}  & \textbf{0.95}    &  \textbf{11.12s}   \\ 
\bottomrule
\end{tabular}
\end{table*}

\vspace{-2mm}
\paragraph{Quantitative Comparison} We conduct quantitative analysis and comparison on our model and other state-of-the-art methods. Specifically, we compare our model with Point-UV Diffusion~\cite{yu2023texture} and TexOct~\cite{Liu_2024_CVPR} which also train single-category models on ShapeNet. We also compare with some methods using pre-trained 2D diffusion prior for multi-view images synthesis such as TEXTure~\cite{yu2023texture}, Text2Tex~\cite{chen2023text2tex} Paint3D~\cite{zeng2024paint3d}, and TexPainter~\cite{zhang2024texpainter} whose input text prompts are set to ``a *" and `*' is the name of corresponding category. Due to the relatively long time consumption, we only use $50$ 3D objects selected from the test set to evaluate the methods based on 2D diffusion models. We conduct only a qualitative comparison for TexPainter~\cite{zhang2024texpainter} for its lengthy processing time. Table~\ref{tab:uncond} reports the comparison of FID and KID scores as well as average inference time on a single NVIDIA A100 (40G) GPU. It is worth noting that our model only takes about one second to predict the parameters of each 3D Gaussian and the rest of the time is used for texture baking. From Table~\ref{tab:uncond}, we have the following observation.
First, our method obtains the best performance in terms of FID and KID. For example, our method outperforms TexOct~\cite{Liu_2024_CVPR}, by an average of $9.69$ in FID and $0.53$ in KID. These improvements indicate that our method excels at generating high-quality textures. 
Second, TexGaussian achieves the fastest generative speed in all categories, which is much less time-consuming than other methods.

\begin{figure}[t]
  \centering
  \includegraphics[width=\columnwidth]{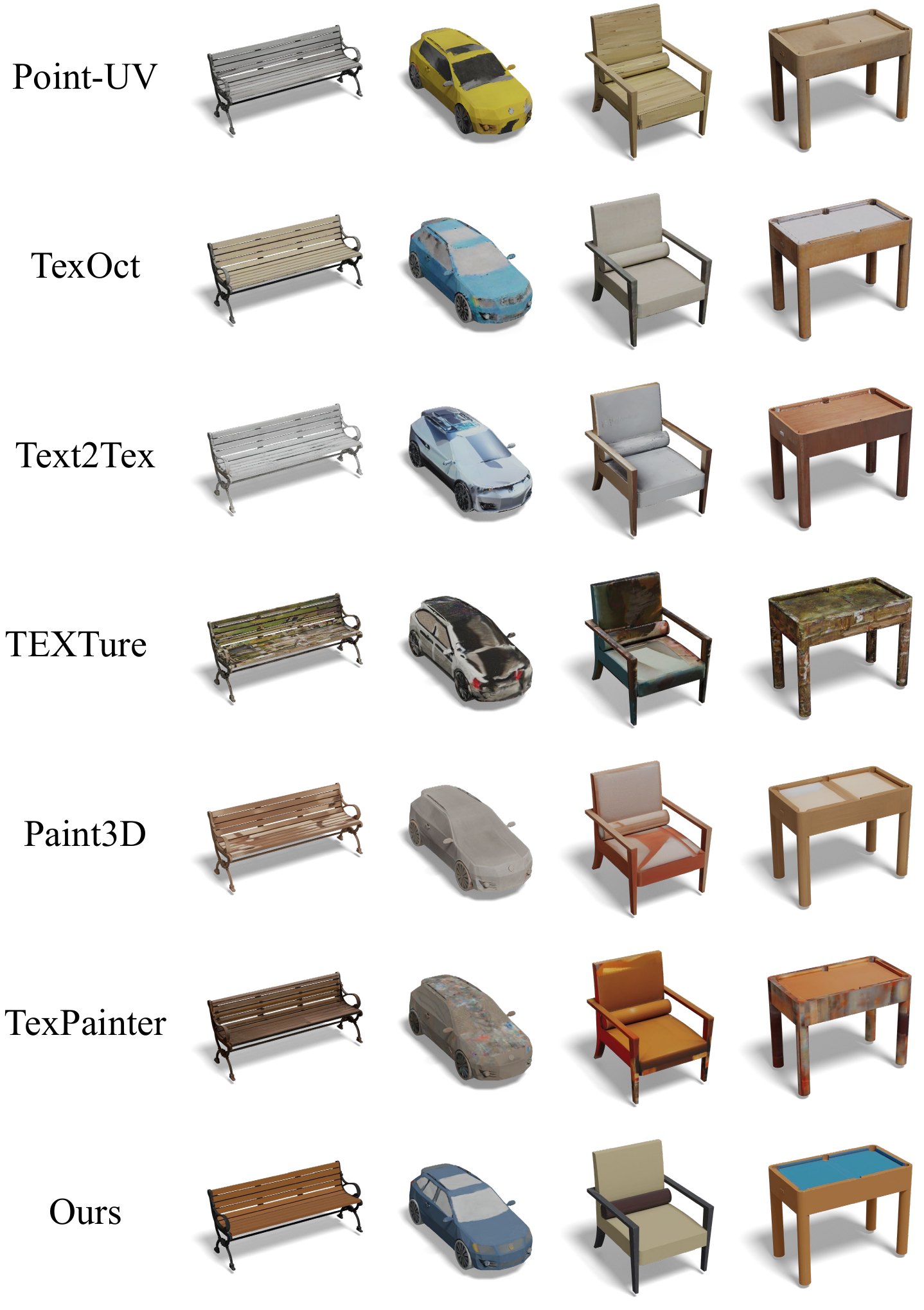}
  \caption{Examples of generated RGB textures obtained by TexGaussian and other state-of-the-art models on the same 3D object. Please zoom in for a better inspection.}
  \vspace{-7mm}
  \label{fig:shapenet_compare}
\end{figure}

\vspace{-3mm}
\paragraph{Qualitative Comparison} 
\cref{fig:shapenet_compare} provides some qualitative results on the same input 3D meshes by different methods. Point-UV~\cite{yu2023texture} and TexOct~\cite{Liu_2024_CVPR} use colored point clouds as supervision to train diffusion models. As a consequence, the generated texture map is relatively blurry due to the discreteness of the point cloud. For other methods that leverage pre-trained 2D diffusion prior, they can hardly comprehend the total geometry. The generated texture maps by them are not consistent with the semantics of input 3D mesh, such as the chaotic stripes on the bench and chair in~\cref{fig:shapenet_compare}. On the contrary, our method generates smooth and colorful RGB textures on unseen objects, which align well with 3D meshes.

% Texturify~\cite{siddiqui2022texturify} employs face convolutional operators on mesh surface and is trained end-to-end using a non-saturating GAN loss. It can only perform convolution operations on watertight meshes which strict greatly limits its applications. 

\begin{figure*}
  \centering
  \includegraphics[width=\linewidth]{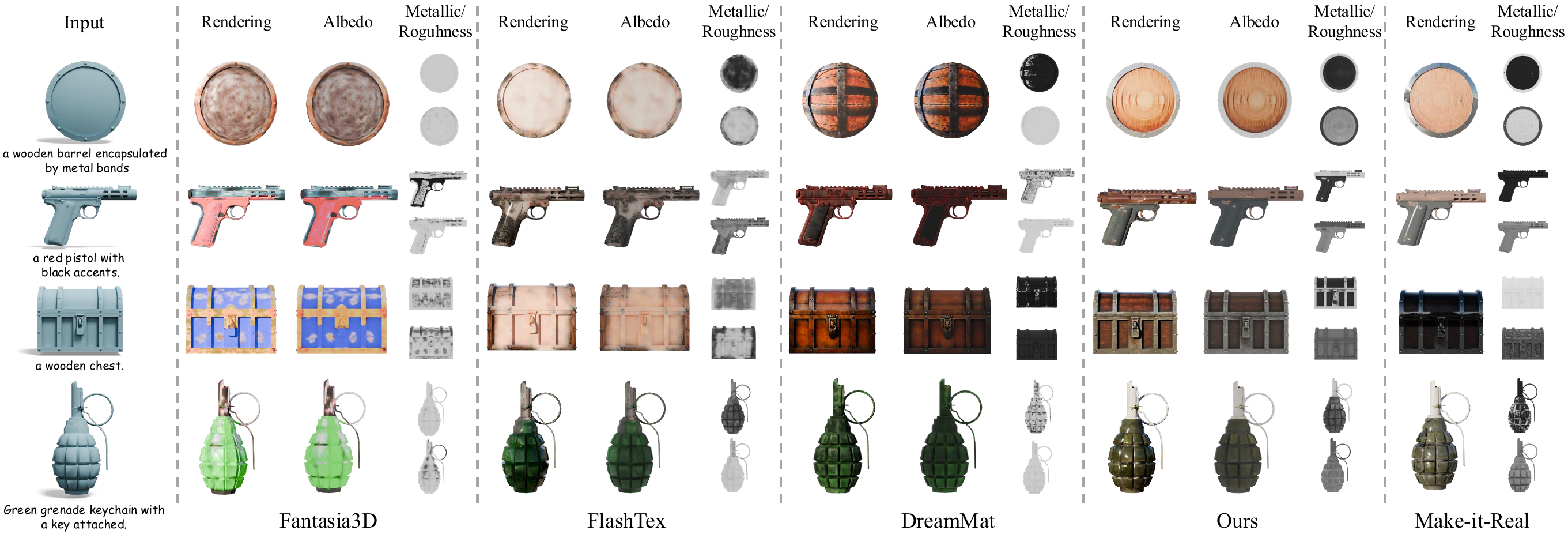}
  \caption{Qualitative comparison with Fantasia3D~\cite{chen2023fantasia3d}, FlashTex~\cite{deng2024flashtex}, DreamMat~\cite{zhang2024dreammat} and Make-it-Real~\cite{fang2024makeitreal}. We show the rendered image, albedo map, roughness map, and metallic map for each 3D object.}
  \label{fig:pbr_compare}
\end{figure*}

\vspace{-2mm}
\subsection{Text-conditioned PBR Material Generation}
We train the text-conditioned PBR material generation model on our filtered subset of Objaverse~\cite{deitke2023objaverse}. We use the text descriptions of 3D objects from Cap3D~\cite{luo2023scalable, luo2024view} to train our model. We choose $29,000$ 3D objects in our filtered subset for training and the rest $200$ for testing. Fig.~\ref{fig:teaser} shows some generative results by proposed TexGaussian on the test set. We can see that our model comprehends the overall geometry feature of input 3D mesh. It is capable of synthesizing high-quality albedo, roughness, and metallic, which are aligned well with 3D semantics such as accents with high metallic on the wooden box and rope with high roughness on the jug.

\vspace{-3mm}
\paragraph{Quantitative Comparison} We conduct quantitative comparison with some state-of-the-art text-conditioned PBR material synthesis methods: Fantasia3D~\cite{chen2023fantasia3d}, FlashTex~\cite{deng2024flashtex}, and DreamMat~\cite{zhang2024dreammat} on our test set. Table~\ref{tab:pbr} reports FID and KID scores on both multi-view albedo map and PBR rendering images under the same illumination environment. TexGaussian outperforms all the baseline methods in achieving the best visual quality of the generated appearances. We also report the average inference time of different methods on a single NVIDIA A100 (40G) GPU across our test set. Due to the iterative optimization of score distillation sampling, all other three methods cost at least 20 minutes for the generation. In contrast, our TexGaussian only takes about 20 seconds, in which one second is for predicting the Gaussian parameters and the rest is for baking, which performs $60\times$ faster than previous methods.

\vspace{-3mm}
\paragraph{Qualitative Comparison} Fig.~\ref{fig:pbr_compare} visualizes the generated albedo, roughness, and metallic of each compared method from the same text prompt and untextured meshes. We also show the PBR rendering images of the generated materials under the same environment light. It can be observed that Fantasia3D~\cite{chen2023fantasia3d} and FlashTex~\cite{deng2024flashtex} generate irregular colors on mesh surfaces, which lead to rendering results with low quality. DreamMat~\cite{zhang2024dreammat} is capable of generating a visually pleasing appearance for test 3D models. However, it can hardly align the generated PBR material well with the 3D semantics, such as the black band on the wooden barrel. What's more, it tends to generate over-saturated colors, which is demonstrated in the wooden chest and green grenade. In addition, all of the compared methods struggle to completely disentangle the light and texture, which results in relatively chaotic results of generated metallic and roughness. We also compare our method with a Multimodal Large Language Model based method, Make-it-Real~\cite{fang2024makeitreal}, which accepts 3D objects paired with albedo maps generated by our TexGaussian as input. It can be observed that Make-it-Real struggles to capture the overall 3D structure of input meshes. As a consequence, it tends to predict single-value PBR maps, leading to irregular rendering results, such as the wooden chest with extremely large metallic value. On the contrary, our method is directly trained from 3D original data. It is capable of generating clean and smooth PBR material while fully comprehending the overall 3D structure. We provide more generative results in the supplementary material.

\begin{table}[t]
\centering
% \tablestyle{7pt}

\setlength{\tabcolsep}{4.5pt}
\small
\caption{Quantitative comparison of the FID and KID ($\times 10^2$) scores as well as the inference time of TexGaussian and other methods on our test set which consists of 200 3D objects with ground truth PBR materials.}
\label{tab:pbr}
\begin{tabular}{lccccc}
\toprule
\multirow{2}{*}{Methods}                                   & \multicolumn{2}{c}{Albedo}        & \multicolumn{2}{c}{PBR rendering} & \multirow{2}{*}{Time} \\ \cline{2-5}
                                                           & FID$\downarrow$ & KID$\downarrow$ & FID$\downarrow$ & KID$\downarrow$ &                       \\ \hline
Fantasia3D~\cite{chen2023fantasia3d} & 213.21          & 0.96            & 209.87          & 0.48            & 22.7mins              \\
FlashTex~\cite{deng2024flashtex}     & 185.24          & 1.13            & 186.82          & 0.42            & 20.3mins              \\
DreamMat~\cite{zhang2024dreammat}    & 152.63          & 1.09            & 145.49          & 0.19            & 48.2mins                \\
Ours                                                       & \textbf{123.72 }         & \textbf{0.20 }           & \textbf{129.52 }         & \textbf{0.02 }           & \textbf{21.04s}                \\ \bottomrule
\end{tabular}
% \vspace{2cm}
\vspace{-3mm}
\end{table}

% \vspace{-2mm}
\subsection{Ablation Study}
% For the purpose of analyzing the impact of different designs in our model, we conduct ablation studies by removing or changing some proposed modules.

\begin{table}[h]
  % \tablestyle{4pt}{1.2}
  \setlength{\tabcolsep}{0.6pt}
  \small
  \caption{Quantitative comparison of methods using different types of 3D Gaussian Fitting.}
  \begin{tabular}{lcccc}
    \toprule
    Method & PSNR $\uparrow$ & LPIPS $\downarrow$ & SSIM $\uparrow$ & 3D Gaussian Number \\
    \midrule
    Full voxel aligned & 23.75 & 0.17 & 0.92 & 16,777,216 ($256^3$) \\
    Octant-aligned & \textbf{32.60} & \textbf{0.039} & \textbf{0.97} & \textbf{79,480} \\
    \bottomrule
  \end{tabular}
  \vspace{-0.3cm}
  \label{tab:ablation}
\end{table}

\begin{figure}[t]
% \vspace{-3mm}
  \centering
  \includegraphics[width=\columnwidth]{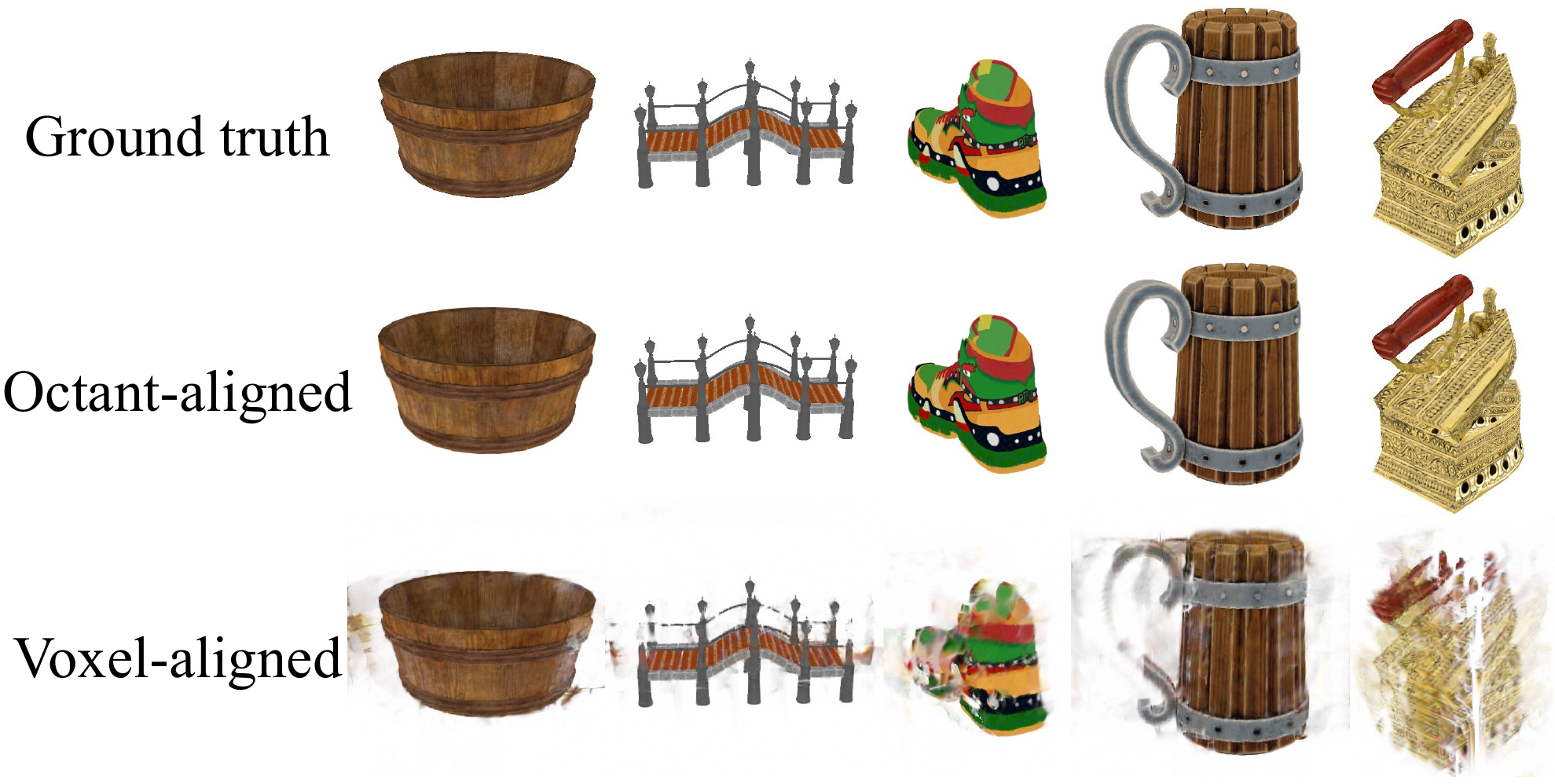}
  \caption{Visualization of different manners of Gaussian fitting. The rendering results demonstrate excellent reconstruction performance of the proposed octant-aligned 3D Gaussian Splatting.}
  \vspace{-5mm}
  \label{fig:recon}
\end{figure}

% \vspace{-2mm}
\paragraph{Essentials of Octree} We first analyze the effectiveness of using Octree in our pipeline. To do this, we calculate the quantitative quality of 3D Gaussian pre-fitting by our octant-aligned and full voxel-aligned 3D Gaussian Splatting on the subset of Objaverse dataset~\cite{deitke2023objaverse}. The depth of octree in this ablation is set to 8, and the resolution of full voxels is $256^3$, which is consistent with the finest resolution of octree to guarantee comparison fairness. Table~\ref{tab:ablation} reports the PSNR, LPIPS~\cite{zhang2018unreasonable}, and SSIM~\cite{wang2004image} metrics on the albedo map as well as the average number of 3D Gaussians of octant and full voxel aligned 3D Gaussian Splatting across our Objaverse subset. \cref{fig:recon} also presents some visual results of our Gaussian fitting. These results demonstrate that the proposed octant-aligned 3D Gaussian produces a much more reasonably compact and precise representation of diverse and complex 3D assets with much fewer 3D Gaussians compared to the full voxel version.

\begin{figure}[t!]
  \centering
  \includegraphics[width=0.95\linewidth]{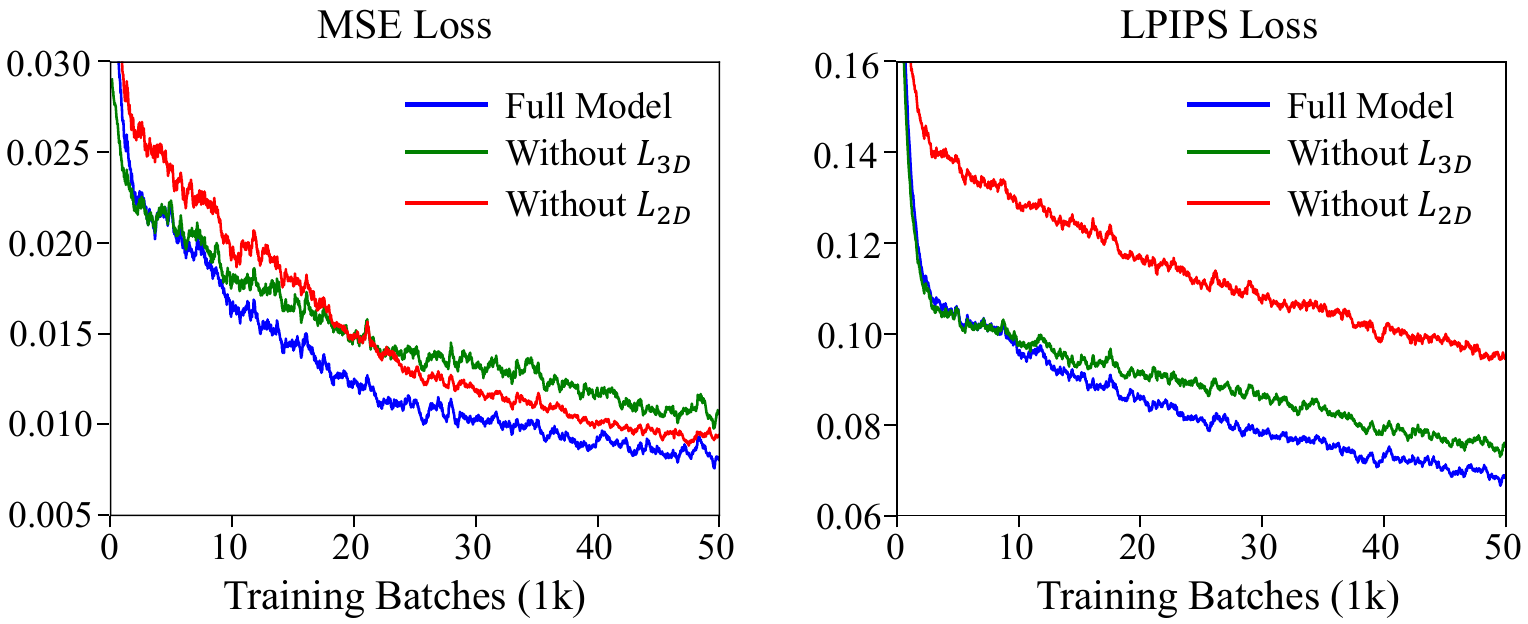}
  \caption{The training loss curves of our model with different losses.}
  \label{fig:loss}
\end{figure}

\begin{figure}[t]
  \centering
  \includegraphics[width=\columnwidth]{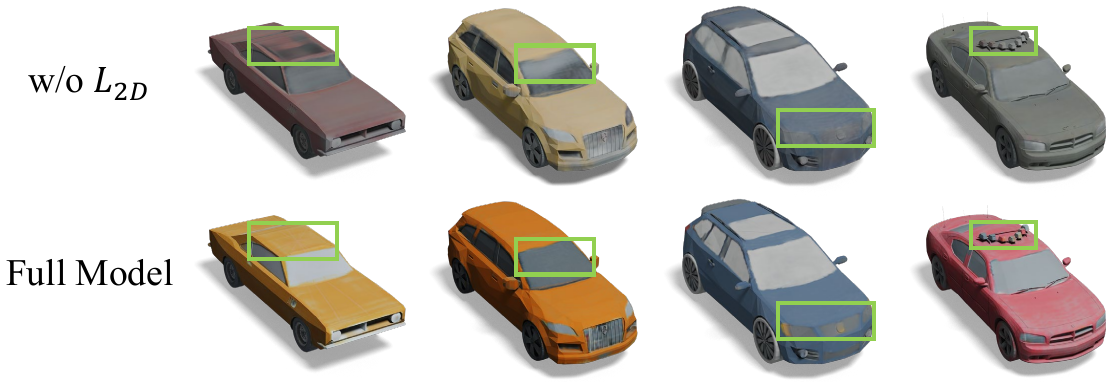}
  \caption{Results of our model trained with and without $L_{2D}$.}
  \label{fig:ablation}
\end{figure}

\vspace{-5mm}
\paragraph{Essentials of different losses} We verify the importance of different losses we proposed to train our model. We term all the losses calculated on 2D image space as $L_\text{2D}$:
\begin{equation}
    L_\text{2D} = L_\text{albedo} + L_\text{R} + L_\text{M} + L_\alpha.
\end{equation}
We train two additional TexGaussian models using only the $L_\text{2D}$ or $L_\text{3D}$ loss on ShapeNet \texttt{car} category due to its large variations and complexity of texture to validate their effectiveness. It is worth noting that in this dataset, $L_\text{2D} = L_\text{albedo} + L_\alpha$ due to the lack of material information. The training curves of MSE loss and LPIPS loss between rendering images and ground-truth ones are shown in Fig.~\ref{fig:loss}, which demonstrate the effects of $L_{2D}$ and $L_{3D}$. From the loss curves, we can conclude that $L_\text{3D}$ facilitates the process of convergence and $L_{2D}$ enhances the quality of synthesized images, as evidenced by the large margin of improvement in the LPIPS loss when introducing $L_{2D}$. We also provide some qualitative results to verify the effectiveness of $L_{2D}$ on the test set of \texttt{car} category in Fig.~\ref{fig:ablation}. Only using $L_{3D}$ results in a relatively blurry texture map due to the discreteness of 3D Gaussian which is similar to the characteristic of 3D point clouds analyzed above.

\section{Conclusion}
In this paper, we proposed TexGaussian, an octree-based 3D Gaussian Splatting model for high-quality PBR material generation on untextured meshes. We aligned each 3D Gaussian on the octant of the corresponding octree built from the input untextured object and extended the parameters of 3D Gaussian with additional channels to represent the roughness and metallic map. We trained our model with regression objectives, achieving faster inference speed compared to previous texture synthesis methods. Experimental results demonstrated that our method is capable of generating high-quality PBR materials that are readily usable in modern graphics engines for photo-realistic rendering, offering enhanced realism for a variety of applications.

\vspace{-2mm}
\paragraph{Limitations} The generalization of TexGaussian is still hindered by the scale of the training set. Thus it struggles to generate various textures for some extremely complex 3D objects beyond our training data. We are looking forward to training our TexGaussian model with more parameters and more data on a larger-scale GPU cluster in the future.

\small
\bibliographystyle{ieeenat_fullname}
\bibliography{egbib}

\clearpage
\appendix

%%%%%%%%% BODY TEXT
\section{Preliminary of 3D Gaussian Splatting}

Gaussian splatting employs a collection of 3D Gaussians to represent 3D data. Specifically, each Gaussian is formally defined as:
\begin{equation}
    G(\boldsymbol{x})=e^{-\frac{1}{2}(\boldsymbol{x}-\boldsymbol{\mu})^T \boldsymbol\Sigma^{-1}(\boldsymbol{x}-\boldsymbol{\mu})},
\end{equation}
where $\boldsymbol{\mu} \in \mathbb{R}^3$ represents the spatial mean of 3D Gaussian and $\boldsymbol{\Sigma} \in \mathbb{R}^{3\times 3}$ denotes the covariance matrix. The covariance matrix $\boldsymbol{\Sigma}$ of a 3D Gaussian is analogous to describing the configuration of an ellipsoid. Thus, the covariance matrix $\boldsymbol{\Sigma}$ is decomposed into a scaling matrix $\boldsymbol{S}$ and a rotation matrix $\boldsymbol{R}$ as follows:
\begin{equation}
\boldsymbol{\Sigma}=\boldsymbol{R S S}^{\top} \boldsymbol{R}^{\top}.
\end{equation}
To allow independent optimization of both factors, they are stored  separately: a 3D vector $\boldsymbol{s}$ for scaling and a quaternion $\boldsymbol{q}$ to represent rotation.
During the rendering process, the 3D Gaussians are projected onto a 2D plane. With the intrinsic matrix $\boldsymbol{K}$ and extrinsic matrix $\boldsymbol{T}$, the 2D mean $\boldsymbol{\mu}'$ and covariance $\boldsymbol{\Sigma}'$ are defined as follows:
\begin{equation}
\boldsymbol{\mu}^{\prime}=\boldsymbol{K}[\boldsymbol{\mu}, 1]^{\top}, \quad \boldsymbol\Sigma^{\prime}=\boldsymbol{J} \boldsymbol{T} \boldsymbol\Sigma \boldsymbol{T}^{\top} \boldsymbol{J}^{\top}.
\end{equation}
Here, $\boldsymbol{J}$ represents the Jacobian of the affine approximation of the projective transformation. Each 3D Gaussian is associated with an opacity value $\boldsymbol{o}$ and a view-dependent color $\boldsymbol{c}$, determined by a set of spherical harmonics coefficients. In our model, the multi-view rendered images of albedo map do not depend on the selected viewpoints. As a result, we just use three-channels RGB on each 3D Gaussian to represent the view-independent colors instead of original spherical harmonics, and we exclude the positional parameter $\boldsymbol{\mu}$ because each 3D Gaussian is fixed at the center of each finest leaf node of the constructed octree.
All the parameters can be collectively denoted by ${\Theta_0}$ with:
\begin{equation}
{\Theta_0}_i = \{\boldsymbol{o_i}, \boldsymbol{s_i}, \boldsymbol{q_i}, \boldsymbol{c_i}\},
\end{equation}
representing the parameters for the $i$-th Gaussian.

Moreover, to encode the PBR material parameters, we append additional two parameters: roughness $\boldsymbol{r}$ and metallic $\boldsymbol{m}$ at the end of the original Gaussian parameters. To render multi-view images of these two attributes, we concatenate $\boldsymbol{r}$ and $\boldsymbol{m}$ with previous parameters to obtain:
\begin{equation}
{\Theta_1}_i = \{\boldsymbol{o_i}, \boldsymbol{s_i}, \boldsymbol{q_i}, \boldsymbol{r_i}\},~
{\Theta_2}_i = \{\boldsymbol{o_i}, \boldsymbol{s_i}, \boldsymbol{q_i}, \boldsymbol{m_i}\}.
\end{equation}
Then, all the 3D Gaussians are paired with these two new parameters ${\Theta_1}_i$ and ${\Theta_2}_i$, rendered from multiple viewpoints to get multi-view roughness map and metallic map for further training.

\begin{figure}[t!]
  \centering
  \includegraphics[width=\columnwidth]{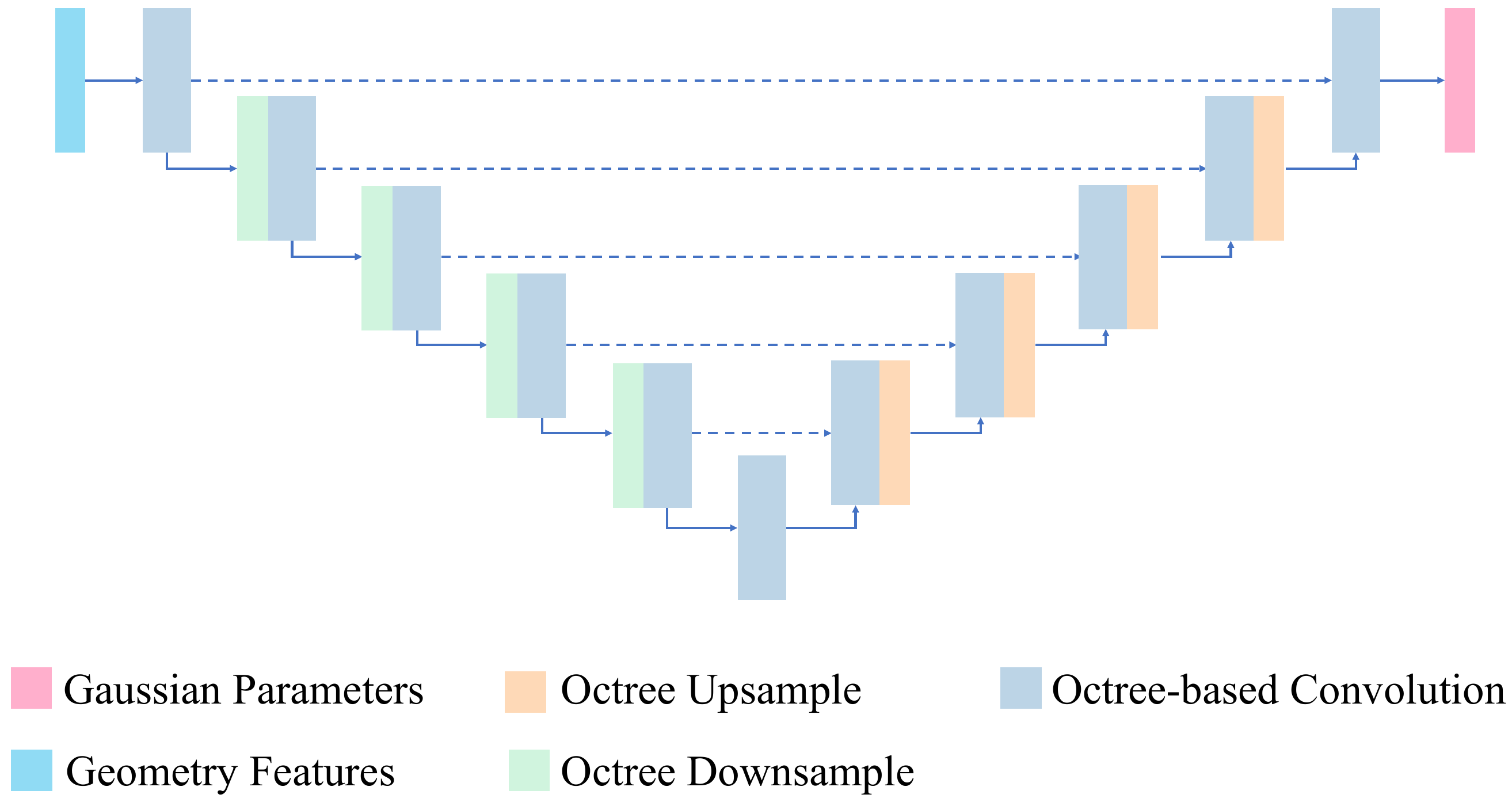}
  \caption{The network architecture of the octree-based 3D U-Net we used to train our unconditional RGB texture generation model.}
  \label{fig:uncond_unet}
\end{figure}

\begin{figure}[t!]
  \centering
  \includegraphics[width=\columnwidth]{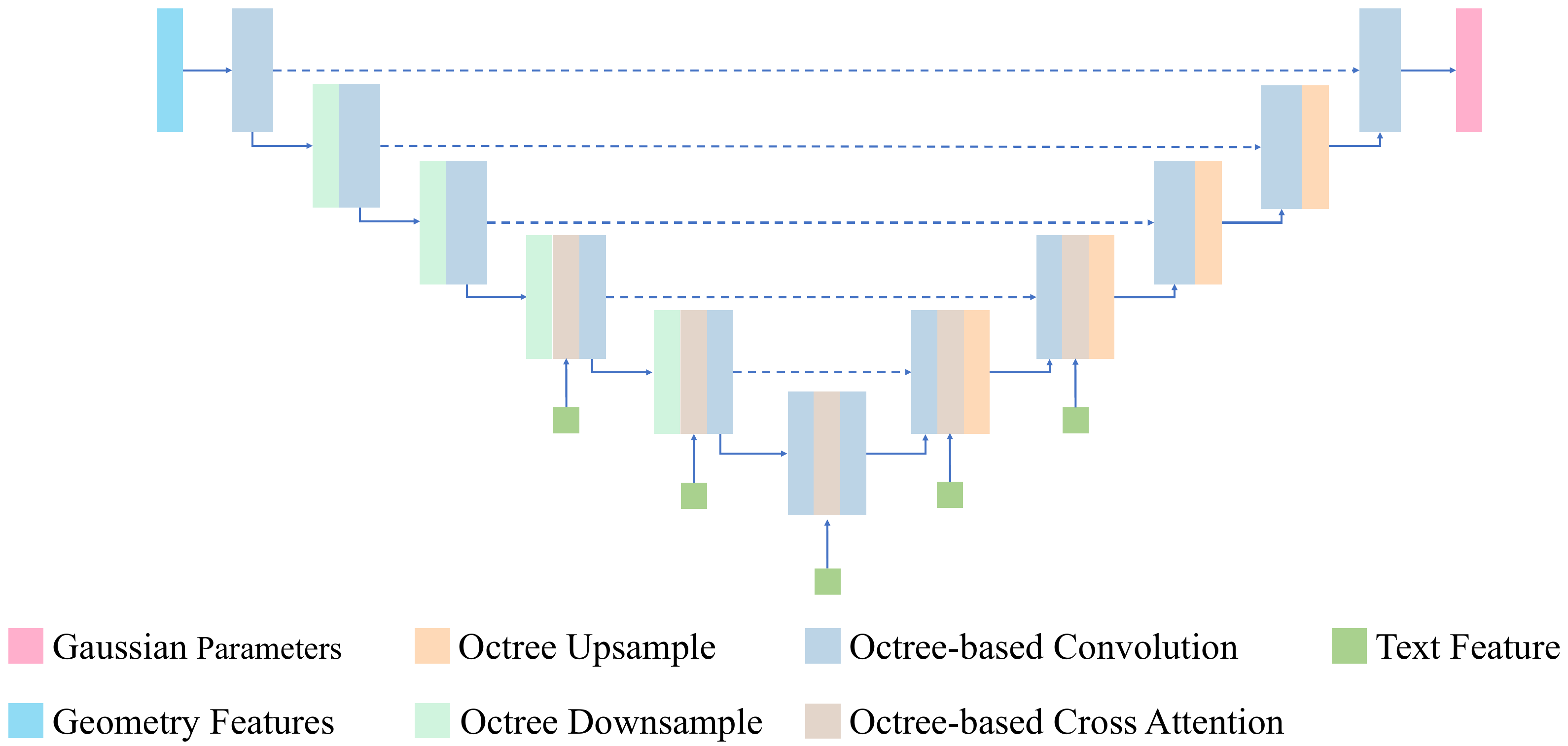}
  \caption{The network architecture of the octree-based 3D U-Net we used to train our text-conditioned PBR material generation model.}
  \label{fig:text_cond_unet}
\end{figure}

\section{Network Details}
\paragraph{Unconditional RGB Texture Generation} The network architecture of the octree-based 3D U-Net we used in unconditional RGB texture generation is shown in Fig.~\ref{fig:uncond_unet}. The U-Net has five hierarchical levels, corresponding to octree depths of 8, 7, 6, 5 and 4, with resolutions of $256^3, 128^3, 64^3, 32^3, 16^3$. The feature dimensions are set to $32, 64, 128, 256, 256$ respectively. The number of channels for input and output features is $4$ and $11$, respectively, due to the lack of material information.

\paragraph{Text-conditioned PBR Material Generation} The network architecture of the octree-based 3D U-Net we used in text-conditioned PBR material generation is shown in Fig.~\ref{fig:text_cond_unet}. The U-Net has five hierarchical levels, corresponding to octree depths of 8, 7, 6, 5 and 4, with resolutions of $256^3, 128^3, 64^3, 32^3, 16^3$. The feature dimensions are set to $64,128, 256, 512, 512$ respectively. The number of channels for input and output features is $4$ and $13$, respectively, as described in the main manuscript. The text feature is fed to U-Net via the octree-based multi-head cross attention mechanism. The cross attention layers are only inserted at the least two down-sampling blocks, the middle block and the two first up-sampling blocks to save GPU memory.

\section{More Results}

\begin{figure*}[t!]
  \centering
  \includegraphics[width=\textwidth]{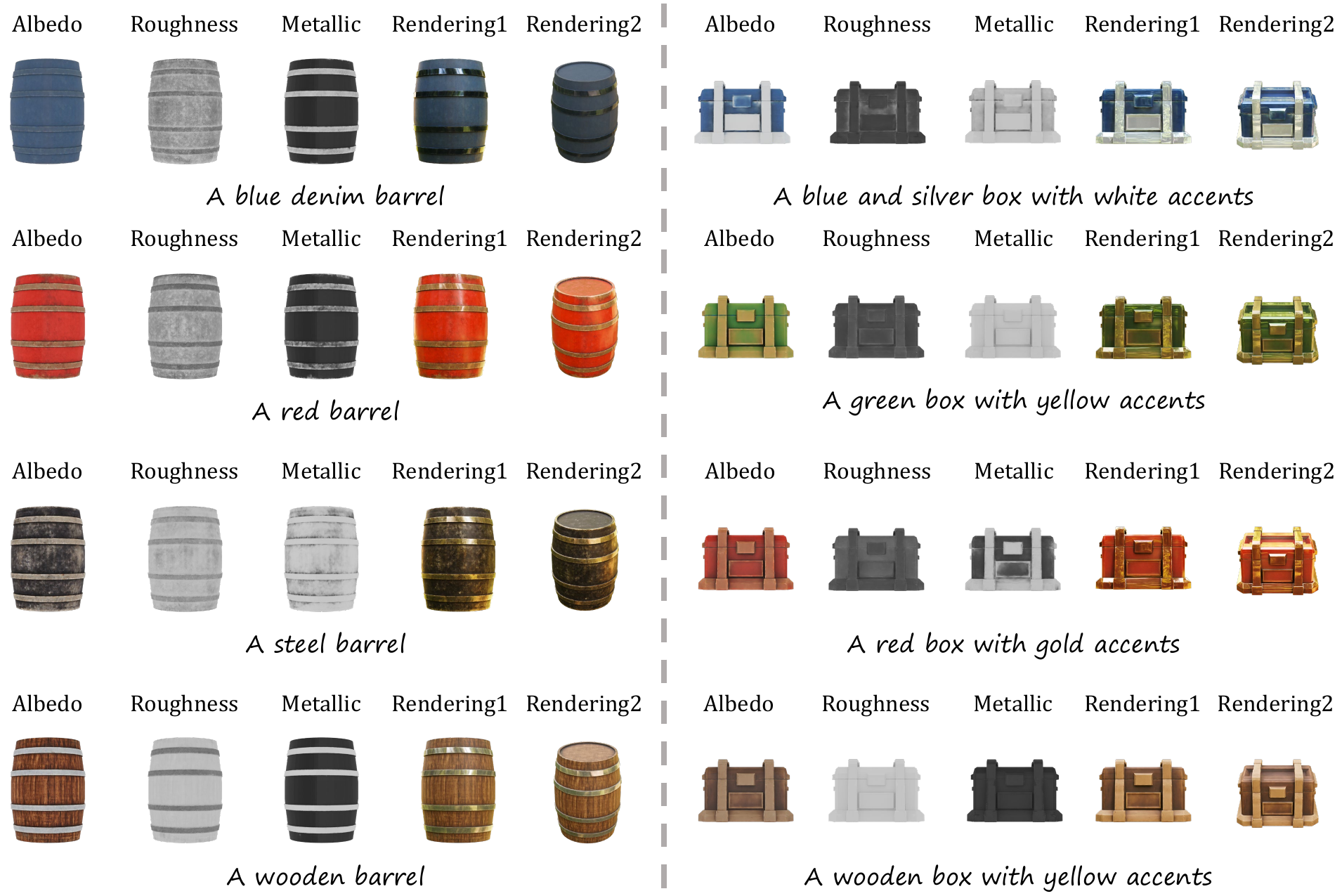}
  \caption{Diverse material generation. Our method can generate different materials with different text prompts on the same mesh.}
  \label{fig:diversity}
\end{figure*}

Our method is capable of generating diverse materials given different text prompts for a single mesh. Fig.~\ref{fig:diversity} shows the PBR materials and the rendering results of the same mesh generated from different text prompts by our proposed TexGaussian. These results demonstrate that our method is able to generate diverse materials of different styles that align well with the text prompts and 3D objects with high fidelity.

\begin{figure*}[t!]
  \centering
  \includegraphics[width=\textwidth]{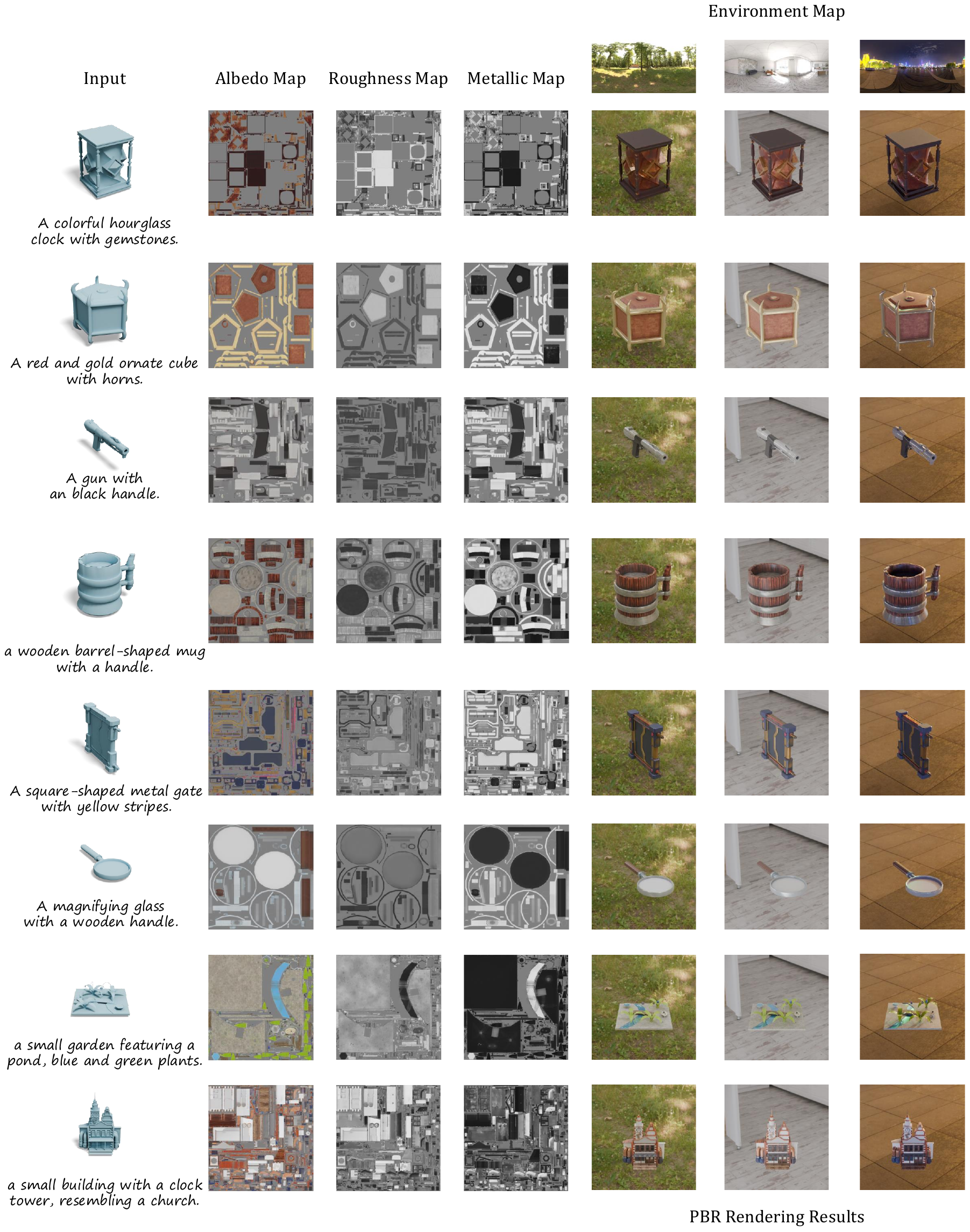}
  \caption{More generative results of our method on different input 3D models and text prompts.}
  \label{fig:suppl_images}
\end{figure*}

We provide more generated results in Fig.~\ref{fig:suppl_images}. 
% Furthermore, we exhibit the \textbf{dynamic rendering effects} of all the 3D objects paired with their generative PBR materials in our paper across different environment maps in \texttt{supplementary\_video.mp4}.

\end{document}